\pdfoutput=1

\documentclass[letterpaper, 10 pt, conference]{ieeeconf}  

\IEEEoverridecommandlockouts                              

\usepackage{multicol}
\usepackage{import}
\usepackage{soul}
\usepackage{wrapfig}
\usepackage{graphicx}
\usepackage{amssymb}
\usepackage{algorithm}
\usepackage[noend]{algpseudocode}

\usepackage{amsmath}

\usepackage{xcolor}
\usepackage[normalem]{ulem}
\usepackage{booktabs}

\renewcommand{\baselinestretch}{.975}

\begin{document}

\title{\LARGE \bf
Preprocessing-based Kinodynamic Motion Planning Framework for Intercepting Projectiles using a Robot Manipulator

}


\author{Ramkumar Natarajan$^{1*}$, Hanlan Yang$^{1*}$, Qintong Xie$^{2}$, Yash Oza$^{1}$, Manash Pratim Das$^{1}$, \\Fahad Islam$^{1}$, Muhammad Suhail Saleem$^{1}$, Howie Choset$^{1}$, and Maxim Likhachev$^{1}$
\thanks{$^{1}$The Robotics Institute, School of Computer Science, Carnegie Mellon University, USA.
        {\tt\small  \{rnataraj, hanlany, yoza, mpratimd, fi, choset, mlikhach\}@andrew.cmu.edu}}%
\thanks{$^{2}$Department of Engineering Science, University of Oxford, UK.
        {\tt\small qintong.xie@univ.ox.ac.uk}}%
\thanks{$^{*}$Contributed Equally
        }%
}

\maketitle
\thispagestyle{empty}
\pagestyle{empty}


\begin{abstract}
We are interested in studying sports with robots and starting with the problem of intercepting a projectile moving toward a robot manipulator equipped with a shield. To successfully perform this task, the robot needs to (i) detect the incoming projectile, (ii) predict the projectile's future motion, (iii) plan a minimum-time rapid trajectory that can evade obstacles and intercept the projectile, and (iv) execute the planned trajectory. These four steps must be performed under the manipulator's dynamic limits and extreme time constraints ($\leq$ 350ms in our setting) to successfully intercept the projectile. In addition, we want these trajectories to be smooth to reduce the robot's joint torques and the impulse on the platform on which it is mounted. To this end, we propose a kinodynamic motion planning framework that preprocesses smooth trajectories offline to allow real-time collision-free executions online. We present an end-to-end pipeline along with our planning framework, including perception, prediction, and execution modules. We evaluate our framework experimentally in simulation and show that it has a higher blocking success rate than the baselines. Further, we deploy our pipeline on a robotic system comprising an industrial arm (ABB IRB-1600) and an onboard stereo camera (ZED 2i), which achieves a 78$\%$ success rate in projectile interceptions.

\end{abstract}

\section{Introduction}
\label{sec:Intro}
\begin{figure*}[]
\centering
 \includegraphics[width=1\textwidth]{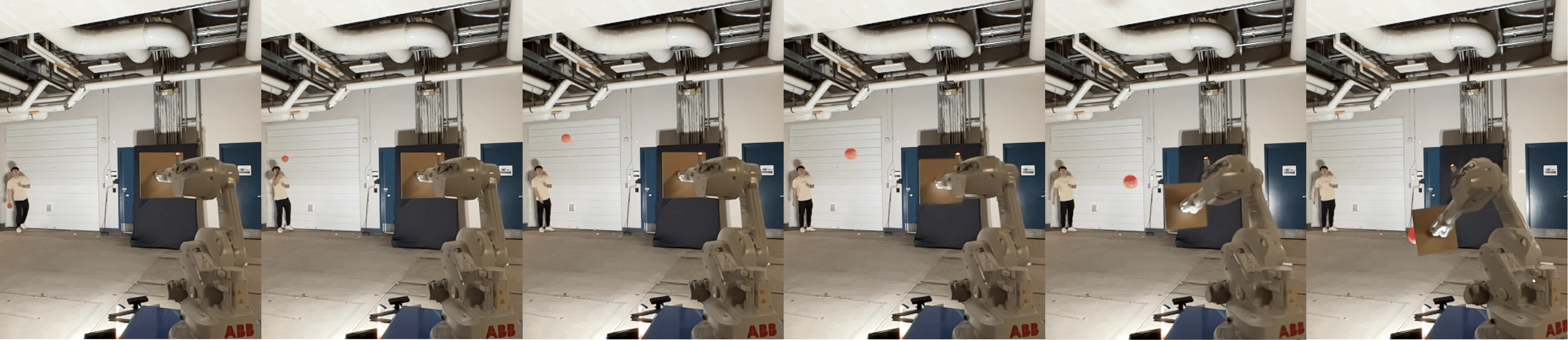}
\caption{ABB arm with a shield attached to its end-effector blocking a ball thrown at it.}
\label{fig:shielding_abb}
\vspace{-0.3cm}
\end{figure*}

In this paper, we study the problem of intercepting projectiles using robot manipulators. This is a challenge that finds unique and exciting applications in various sports. Recent breakthroughs in this field include instances where a robot can handle various types of spin in table tennis matches against human opponents \cite{tebbe2020spin}, and instances where robots have skillfully intercepted flying, rolling, and bouncing balls \cite{7018920}. These scenarios represent the exciting intersection of sports and robotics. In our problem, our primary goal is to intercept incoming projectiles with robot manipulators in real-time with onboard processing. We aim to demonstrate how robotic systems, equipped with kinodynamic planning capabilities, can plan smooth trajectories while minimizing torque requirements.


The time-critical nature of the task calls for an intelligent framework that can maximize the robot’s chances of intercepting projectiles. Based on the range of the vision system and average velocities of the incoming projectiles in our setting, the robot typically has about 350 milliseconds from the time the projectile is first detected until it hits the body of the robot. Within this duration, our framework should be capable of performing three major tasks: (i) detecting and accurately predicting the motion of the incoming projectile, (ii) querying a motion planner for a trajectory that would enable the robot to safely intercept the projectile, and (iii) executing the trajectory returned by the motion planner. The fast nature of the task, combined with the physical limitations of the robot (typical executions of blocking maneuvers consume about 250 of the 350 milliseconds available to us), calls for a real-time, yet optimal, planning framework. In this paper, we present our pipeline that comprises (i) an on-board stereo camera-based perception module that provides estimates of the incoming projectile and its predicted trajectory and (ii) a preprocessing-based planning framework that guarantees to return a blocking trajectory if it exists (based on the perception estimates) within a significantly small time budget (1ms). The contributions of our work can be summarized as follows:  
\begin{itemize}
    \item The formulation of the problem of robot protection against incoming projectiles and the development of the framework consisting of real-time perception, motion planning, and execution.
    \item A preprocessing-based motion planning module that guarantees to return blocking trajectories within an extremely small time window.
    \item A \textit{dome-based} discretization technique that makes preprocessing tractable while still providing strong guarantees.
    \item Demonstration of the effectiveness of our pipeline in the real world by deploying it on an ABB's IRB-1600 robot arm and a ZED 2i stereo camera setup, as shown in Fig. \ref{fig:shielding_abb}.
\end{itemize}

\section{Related Work}
\label{sec:rel_works}

Several works in the literature have investigated tasks that are similar to the task of intercepting projectiles. In this section, we will discuss these works as well as their advantages and disadvantages.
In \cite{5651175}, the authors presented a framework that robot manipulators can use to catch a ball. 
Specifically, the catching problem is formulated as a bilevel optimization problem, which produces a joint trajectory that can successfully complete the catching task. The advantage of using an optimization-based trajectory generation algorithm is that trajectories can be quickly generated in simple environments, even for high-dimensional planning problems. There are three key disadvantages in this work compared to our work: (i) as \cite{5651175} solves an optimization problem online while the projectile is in flight, it is not guaranteed to return a solution within a strict time limit; (ii) \cite{5651175} assumes that there are no obstacles in the workspace; (iii) \cite{5651175} uses a VICON system that provides a near ground-truth pose of the incoming object. Hence, the assumptions of the planning and perception problems in \cite{5651175} are stronger than those of our framework as our pipeline avoids obstacles, performs object pose estimation using RGB-D sensors, and does not require a motion capture system. 
\cite{buchler2020learning} and \cite{mulling2013learning} use reinforcement learning in a model-free setting to learn policies that are capable of generating dynamic strokes for the table tennis robot. 
The main advantage of using machine learning-based methods is that the policies learned by the neural network can generalize to new data points that are similar to previously seen data points. However, these approaches require training a large set of data, especially if the scenes are allowed to have obstacles \cite{kakade2003sample}.

On the perception side, RGB-D cameras are very noisy and thus require special filtering techniques to detect incoming projectiles. Model-specific methods, such as \cite{agarwal2020perch} attempt to fit a known 3D model to the point cloud to efficiently find the location of the projectile. 
However, this assumes that the model of the projectile is known. 
There also exist more generic approaches like \cite{kromer20154d} that do not have such requirements. But these methods require either a laser scanner or a VICON system \cite{pfister2014comparative} to detect the moving object. 
Recently, machine learning-based methods have also gained popularity for object detection \cite{ramik2014machine}. For example, YOLO \cite{redmon2016you} can be utilized for object detection and tracking. However, it cannot detect the fast-moving ball at a distance. Therefore, this method requires substantial data for training, making it challenging to predict their performance on new data points. In our work, we employ a straightforward color detection algorithm to detect the projectile using a single stereo camera. This approach improves the practicality of our framework in real-world scenarios.

\section{Problem Formulation}
\label{sec:sys}

Consider a robot manipulator with a shield rigidly attached to the end-effector of the manipulator. The manipulator is tasked with protecting a specific object.
A projectile is launched in the direction of the object to be protected, and the manipulator must intercept the projectile before it collides with that object. 
The state of the projectile, $\mathcal{\rho}$, is represented as a tuple ($\mathcal{\rho}_p$, $\mathcal{\rho}_v$), where $\mathcal{\rho}_p \in \mathbb{R}^3$ represents the position, and $\mathcal{\rho}_v \in \mathbb{R}^3$ represents the velocity of the projectile. The goal of the problem is to intercept the projectile before it collides with the object which is to be protected $\mathcal{O}$. To solve this problem, we make certain simplifying assumptions:
\begin{itemize}
    \item The manipulator always starts from a ``home" configuration $s_{home}$.
    \item The projectile is launched from within the field of view of the camera, allowing us to get early estimates of its trajectory $\mathcal{\rho}$.
    \item At any point in time, only a single projectile is launched in the direction of $\mathcal{O}$.
\end{itemize}

Let $\mathcal{T}_{t}$ be the time of flight of the projectile, measured from the time it is first observed to when it collides with $\mathcal{O}$. 
Let $\mathcal{T}_d$ be the time duration for the perception module to collect enough frames and finish estimating the trajectory.
Finally, let $\mathcal{T}_{p}$ be the time taken to query the motion planner for a manipulator trajectory, and $\mathcal{T}_e$ be the time taken by the manipulator to execute that trajectory.

In order for the manipulator to successfully intercept the projectile, two conditions need to be satisfied. First, given a long enough time-out, a motion planner must be able to compute a trajectory from $s_{home}$ to a goal configuration which can intercept the projectile. Second, the manipulator also needs to ensure that it reaches the goal configuration before the projectile passes that location (in $\mathbb{R}^3$). Specifically, the following equation needs to be satisfied:  
\begin{equation}
  \label{eq:time_traj}
  \mathcal{T}_t \geq \mathcal{T}_d + \mathcal{T}_p + \mathcal{T}_e
\end{equation}


\section{Method}
\label{sec:method}

\subsection{System Overview}

\begin{figure}[!h]
\centering
\includegraphics[width=1.0\columnwidth]{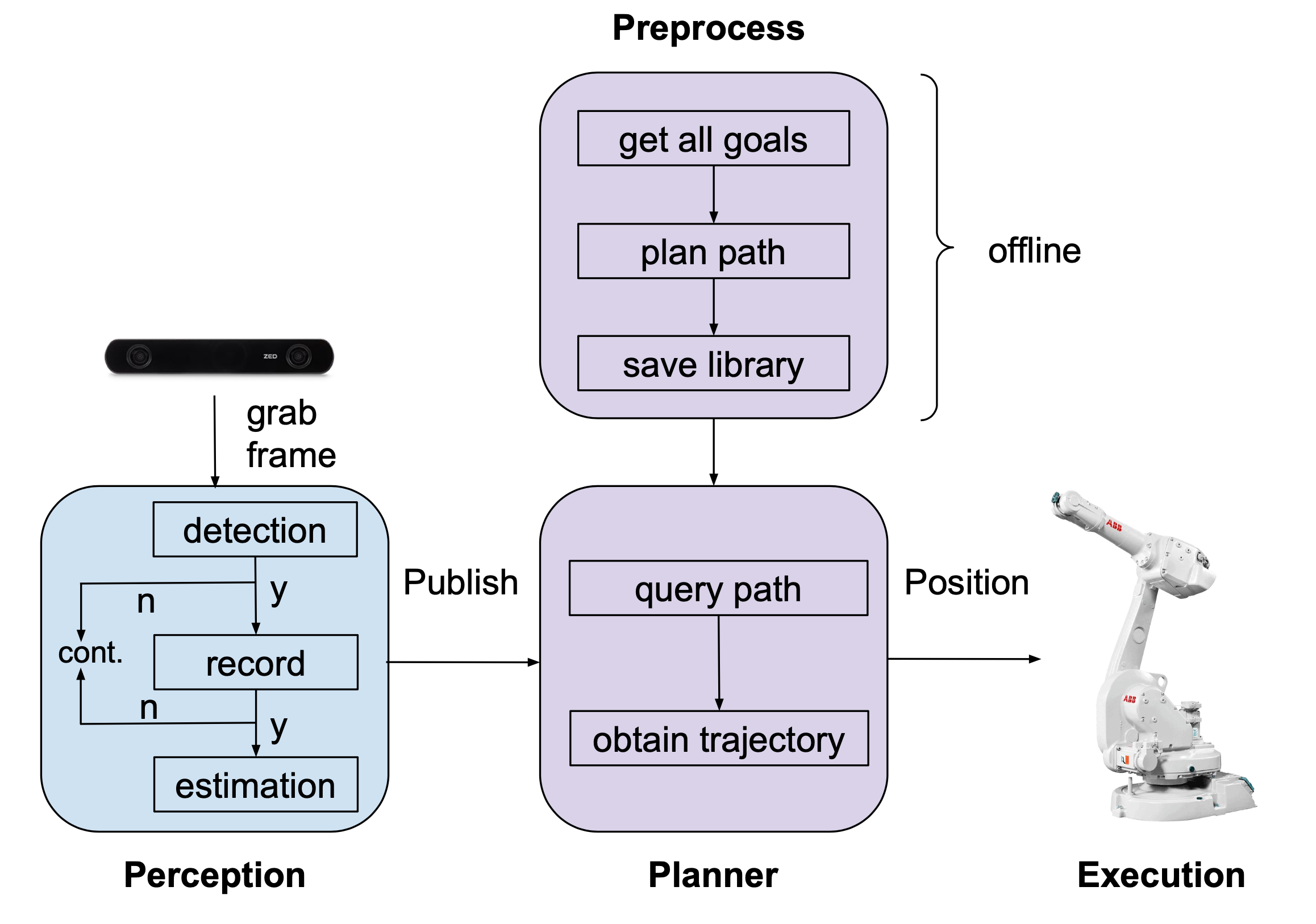}
\caption{Flowchart showing the overall pipeline.}
\label{fig:flowchart}
\end{figure}

The constraints laid out by the problem call for deriving an approach that minimizes time spent on any online operation, hence budgeting as much time as possible for execution and perception. The extreme time constraints limit the use of online real-time planning methods \cite{yang2023epa}. To this end, we propose an approach that relies heavily on precomputation, similar to the Constant-Time Motion Planning (CTMP) class of algorithms \cite{doi:10.1177/02783649211027194}.

At a high level, each frame captured by the camera is processed within the perception module shown in Figure \ref{fig:flowchart}. Once a sufficient number of frames are detected, those frames and their timestamps are used to estimate the trajectory of the projectile. The derived positions and velocities of the projectile are then published to the planner. Subsequently, by querying a motion library computed offline from solutions to a series of planning problems, we obtain a trajectory for the manipulator to block the projectile. The trajectory is then executed to intercept the projectile with the shield held by the manipulator and prevent any potential collisions with the object $\mathcal{O}$. In the remainder of the section, we explain the proposed approach and its building blocks in detail and describe the whole algorithm.

\subsection{Proposed Approach}
\label{sec:prop}
In our approach, we define two cuboidal \emph{domes} around the object $\mathcal{O}$, an inner dome $D_i$ and an outer dome $D_o$. $D_i$ approximates the geometry of the object $\mathcal{O}$ and $D_o$ captures the robot's reachable space so that it can intercept the projectiles with the shield $\mathcal{S}$ positioned anywhere in the 3D space between $D_i$ and $D_o$. The two domes are discretized into cells. Our planning approach is divided into two stages, the preprocessing stage and the query stage. In the preprocessing stage, for each pair of cells (with one cell from $D_i$ and one cell from $D_o$), we plan a path to a pose of $\mathcal{S}$ that can block all projectiles passing through the pair of cells. These paths are stored in a lookup table mapping the pair of cells to the corresponding path. In the query stage, for an incoming projectile $\rho$, we first identify the pair of cells through which $\rho$ passes. Second, we look up the corresponding path $\pi$ from the look-up table in constant time. With this approach, the size of the goal region $\mathcal{G}$ becomes equal to the total number of pairs of cells. Note that these cells are computed only with two-dimensional discretization of the domes' surfaces as opposed to six-dimensional discretization of the space of projectiles. This greatly reduces the size of $\mathcal{G}$.


\subsection{The Building Blocks}
\label{sec:propbb}
\subsubsection{Dome Specifications}
\begin{figure}[!htb]
    \centering
    \begin{tabular}{ccc}
        \includegraphics[width=0.14\textwidth, height=0.9in]{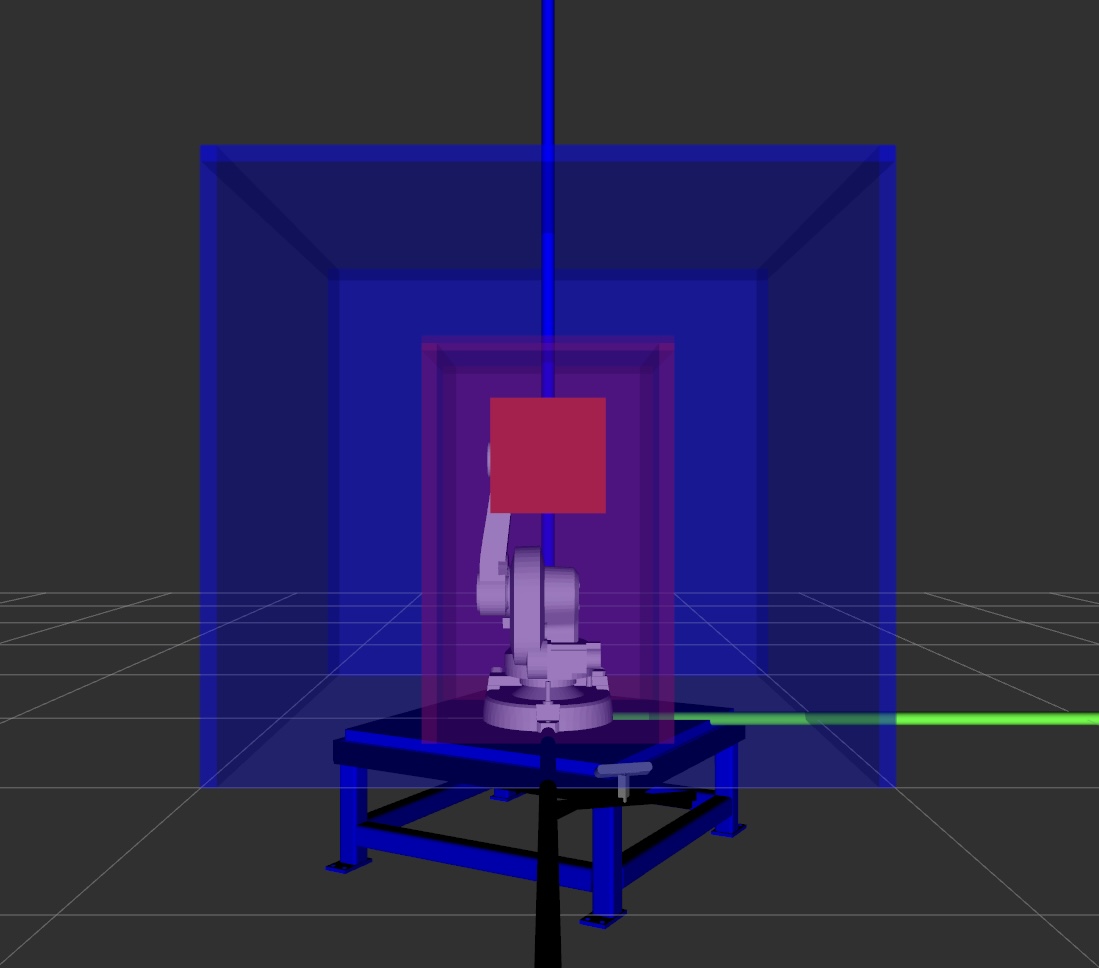} &
        \includegraphics[width=0.14\textwidth, height=0.9in]{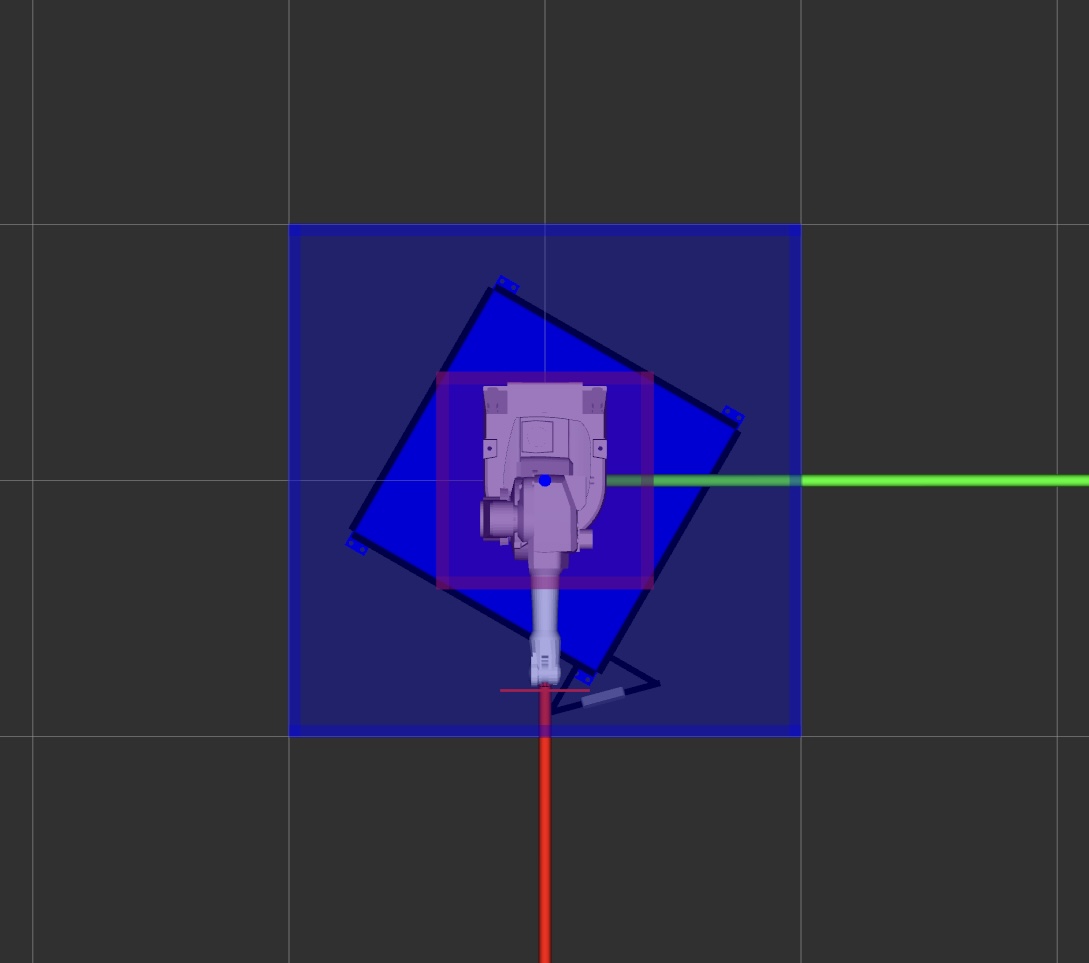} &
        \includegraphics[width=0.14\textwidth, height=0.9in]{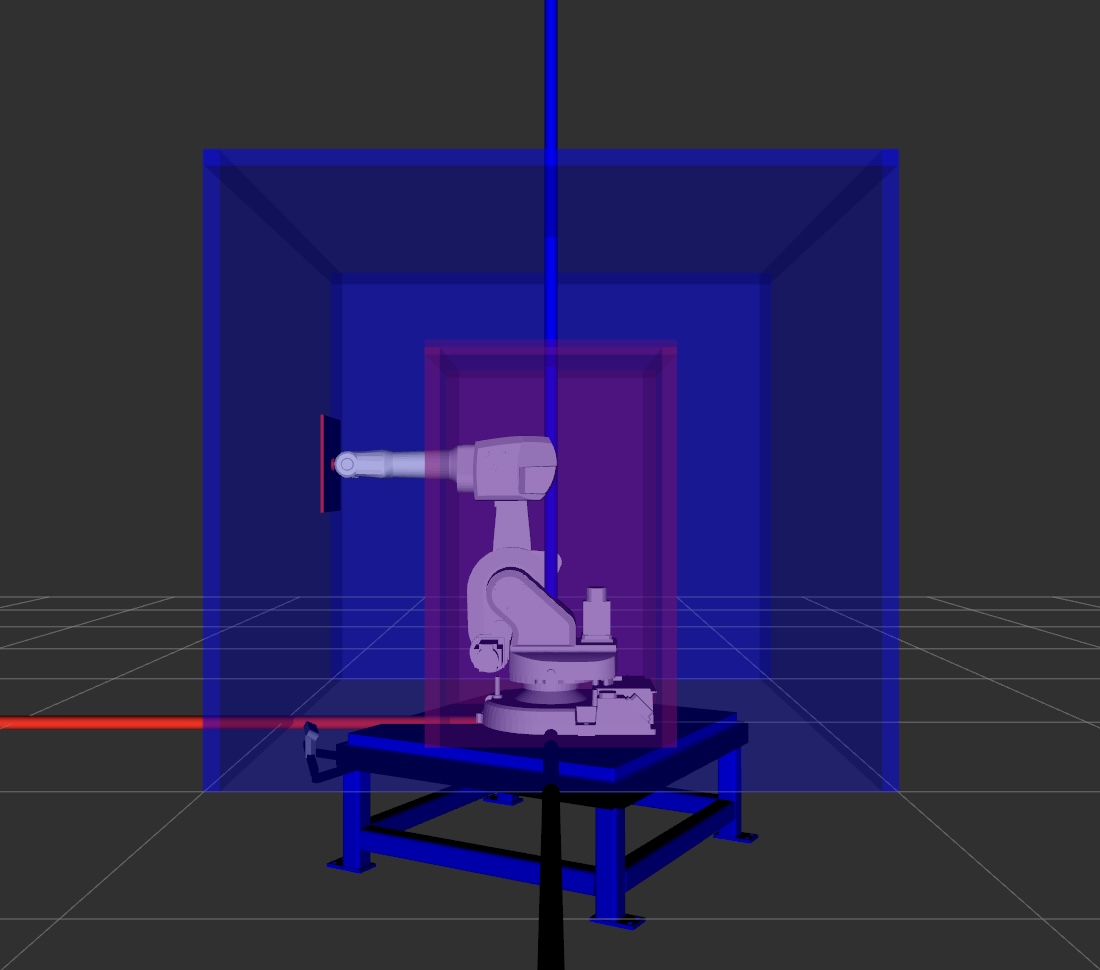} \\ 
\footnotesize{(a) Front view}
\label{fig:domes_front} &
\footnotesize{(b) Top view} 
\label{fig:domes_top}&
\footnotesize{(c) Side view} 
\label{fig:domes_top} \\
        \end{tabular}
        \caption{\footnotesize{(a) and (b) show the inner (red) and outer (blue) domes surrounding the robot. (c) shows the ABB's robot arm with a shield attached to its end-effector (in simulation).}}
        \label{arm_main}
        \vspace{-2mm}
\label{fig:domes}
\end{figure}

The geometry of $D_i$ is such that it tightly encapsulates the object $\mathcal{O}$. In other words, it overestimates the geometry of $\mathcal{O}$ with a simple shape. With this, we simplify the problem setup by making a more conservative requirement that $D_i$ must be protected instead of $\mathcal{O}$. We define $D_i$ as a cuboid. The outer dome $D_o$ captures the reachable workspace of the robot. For simplicity, we also define $D_o$ as a cuboid and place it concentric with $D_i$. A larger $D_o$ would allow more freedom for the robot, but would also increase the demand on preprocessing. On the other hand, a smaller $D_o$ would restrict the robot and limit its protection capability. We choose the size of $D_o$ such that the manipulator can reach the side it faces at full extension. Although our approach is simple, it can easily be generalized to different robot models. Fig. \ref{fig:domes} shows the two domes configured for the ABB robot arm. The volume between $D_o$ and $D_i$ is where the robot manipulates the shield to intercept the incoming projectiles.

\subsubsection{Dome Discretization and Shield Geometry}
\label{sec:cell_size}
Each side of $D_i$ and $D_o$ is discretized into cells. The discretization is correlated with the shape and size of $\mathcal{S}$. We use a square-shaped $\mathcal{S}$ in our setup. The discretization of the two domes is shown in Figure 4 (a). It should also be noted that the size of the shield in our application can be made significantly larger to decrease the preprocessing effort. But doing this is not always practical and could violate the task and environment constraints. The volume around the straight line connecting a cell from $D_i$ to a cell from $D_o$ constitutes a \emph{tunnel}. A line segment connecting the centers of the cells that form the tunnel is called \emph{centerline}. Our key idea is that if $\mathcal{S}$ is positioned such that it fully blocks this tunnel, all possible attacks that cross the cell pair are blocked by it. This idea is illustrated in Figure 4 (b).

\begin{figure}[!htb]
    \centering
    \begin{tabular}{cc}
         \includegraphics[width=0.15\textwidth, height=1.0in]{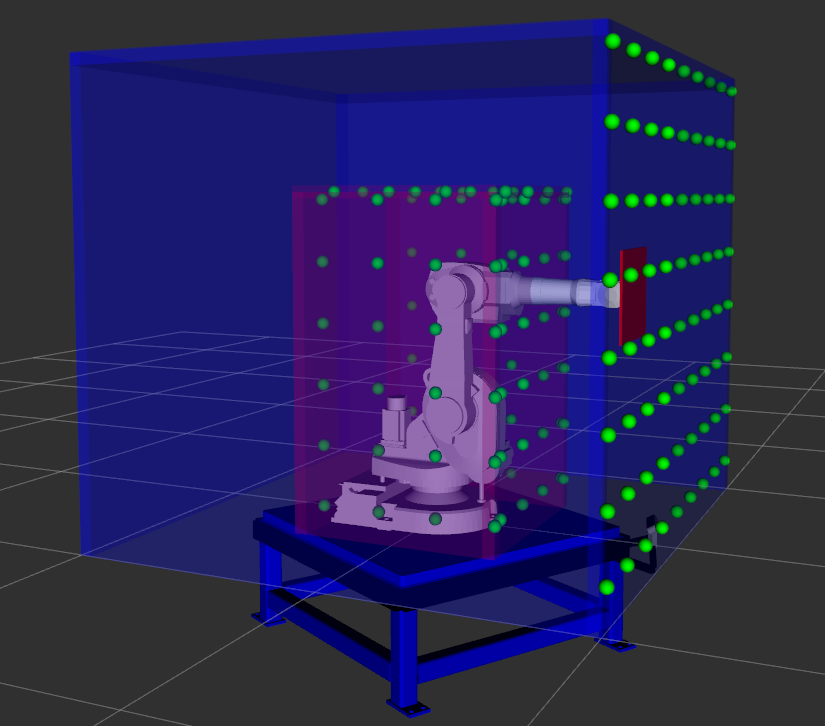} &
        \includegraphics[width=0.2\textwidth, height=1.0in]{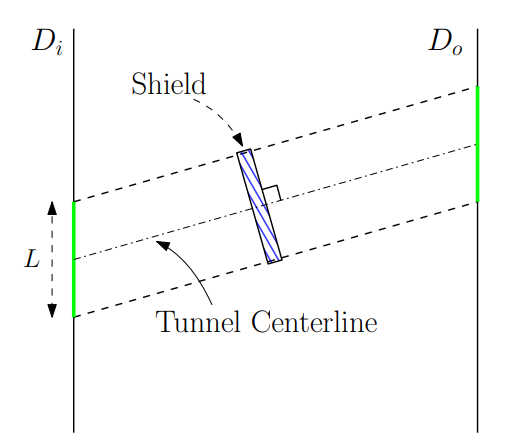} \\ 
\footnotesize{(a) Dome discretization}
\label{fig:shield_idea} &
\footnotesize{(b) Tunnel diagram} 
\label{fig:domes_disc} \\
        \end{tabular}
        \caption{\footnotesize{(a) shows the centers of the cells on both domes. For $D_o$ we only show the discretization for the front side. (b) shows a tunnel formed by a pair of cells (shown in green) in $D_o$ and $D_i$. }}
        \label{domes_main}
        \vspace{-2mm}
\label{fig:tunnel}
\end{figure}

The size of the cell is proportional to the size of $\mathcal{S}$. Specifically, we choose the cell size to be smaller than the size of $\mathcal{S}$ to allow some tolerance in the pose of $\mathcal{S}$ that blocks the tunnel. This tolerance is also needed for possible planning and execution errors. We performed a thorough geometric analysis of the magnitude of the reduction in cell size that is needed to account for these errors. These analyses have been omitted for brevity. 

We approximate the portion of the projectile that lies between the two domes by a line segment. This approximation is made under the assumption that the objects move in a straight line within that region and therefore do not breach the boundaries of the tunnel which they enter. This is not a strong assumption if the distance between $D_i$ and $D_o$ is small compared to the distance from which attacks are launched. This assumption can be further relaxed by reducing the cell size needed to account for the projectile shape to line-segment approximation error. We leave this analysis for future work.

\begin{algorithm}
\label{alg:gen_base_traj}
\begin{footnotesize}
\caption{Generate Trajectory Database}
    \begin{algorithmic}[1]
    \State \textbf{Inputs: } Motion planner $\textbf{P}$, Home configuration $\textbf{x}^{S}$, Dome configuration $\textbf{C}$
    \State \textbf{Output: } Trajectory database $\textbf{D}$
    \Procedure{Preprocess($\textbf{P}, \textbf{x}^{S}, \textbf{C}$)}{}
        \State $\textbf{L} \leftarrow \textproc{ComputeLineSegments}(\textbf{C})$ 
        \For{$l_i$ in $\textbf{L}$}
            \State $\textbf{T}_{i}$ $\leftarrow$ \textproc{ComputeTargetPoses($l_i$)}
            
            \State $TrajBuffer$ = []
            \For{$t$ in $\textbf{T}_i$}
                \If{\textproc{Ik($\textbf{x}^{S}, t$)} succeeds}
                    \If{$\textbf{P}(\textbf{x}^{S}, t)$ succeeds}
                        \State Add resulting plan to $TrajBuffer$.
                    \EndIf
                \EndIf
            \EndFor
            \State Add least-time trajectory in $TrajBuffer$ to $\textbf{D}$.
        \EndFor
    \EndProcedure
    \end{algorithmic}
\end{footnotesize}
\end{algorithm}

\subsubsection{Goal Condition}
The goal $g \in \mathcal{G}$ is defined as the centerline of a tunnel. For the motion planner, the goal condition is any pose of $\mathcal{S}$ along the centerline such that $\mathcal{S}$ is oriented orthogonal to it (see Figure 4 (b)). 
For ease of planning, we allow a small tolerance in $SE(3)$ for the goal pose. In our implementation, we sample equidistant points along the tunnel's centerline and compute $SE(3)$ poses at each point which are orthogonal to it. The motion planner then attempts to plan to each of these poses sequentially until it succeeds. If it fails to do so, then the corresponding $\mathcal{G}$ is marked as unreachable.
 
\subsubsection{Motion Planner}
We use a hybrid kinodynamic motion planner called INSAT (INterleaved Search and Trajectory optimization) that combines the benefits of heuristic search and trajectory optimization. We use a heuristic search-based planning approach with motion primitives (see, e.g.,~\cite{likhachev2003ara}, \cite{likhachev2009planning}, \cite{pohl1970heuristic}) because they have strong theoretical properties. To keep motion planning efficient and tractable, INSAT \cite{insat1, insat2, troptc} interleaves heuristic search in the low-dimensional (low-D) subspace with trajectory optimization in the full-dimensional (full-D) planning space. In our setup, the low-D subspace is the configuration space of the arm. Thus, the states and the transitions in low-D implicitly define a graph $G = (S,E)$ where $S$ is the set of all states and~$E$ is the set of all transitions defined by the motion primitives.
\subsection{Offline Preprocessing and Online Querying}
\label{sec:algo}

\subsubsection{Preprocessing Stage}
In the preprocessing stage, for each pair of cells, the tunnel centerlines are computed. Each tunnel is checked for feasibility and the tunnels whose volume snaps to zero anywhere along the length are discarded because no incoming object coming through such a tunnel can reach the tunnel's end on $D_i$. The centerlines of all the feasible tunnels constitute the goal region $\mathcal{G}$. We pick a constant number of equidistant goals on the line segment to plan for the manipulator ($\textproc{computeTargetPoses}$ method on line 6 in Algorithm 1). Our algorithm computes the paths from $\textbf{x}^{S}$ to cover $\mathcal{G}$ by satisfying the goal criteria described above.

\subsubsection{Query Stage}
In the query stage, the corresponding centerline is identified for a given query $g \in \mathcal{G}$. This process involves first computing the points of intersection of the projectile with $D_o$ and $D_i$ and then identifying the cells containing these points. Subsequently, the associated path is retrieved from the lookup table constructed in the previous step.

\setlength{\textfloatsep}{4pt}
\begin{algorithm}
\begin{footnotesize}
\begin{algorithmic}[1]
\Procedure{Main}{$\textbf{x}^{S}, \textbf{x}^{G}$}
\State $\textbf{x}^{\text{next}}_L = \textbf{x}^S$ 
\State \textbf{while} $\boldsymbol{\phi}_{\textbf{x}^{\text{S}}\textbf{x}^{\text{G}}}$ is EMPTY \textbf{do} \label{lineinsat:termin}
\State $\>$ Pick the next node $\textbf{x}^{\text{next}}_L$ to expand \Comment{\textcolor{blue}{Low-D heuristic search}} \label{lineinsat:pq}
\State $\>$ Generate the successors $\textbf{X}^{\text{new}}_L$ of $\textbf{x}^{\text{next}}_L$ \Comment{\textcolor{blue}{Low-D heuristic search}} \label{lineinsat:succ}
\State $\>$ \textbf{for} $\textbf{x}^{\text{new}}_L$ in $\textbf{X}^{\text{new}}_L$ \textbf{do}
\State $\>$ $\>$ Get the ancestors $\textbf{X}^{\text{pred}}_L$ of $\textbf{x}^{\text{new}}_L$ \label{lineinsat:ances}
\State $\>$ $\>$ \textbf{for} $\textbf{x}^{\text{pred}}_L$ in $\textbf{X}^{\text{pred}}_L$ \textbf{do}
\State $\>$ $\>$ $\>$ $\boldsymbol{\phi_{\textbf{x}^{\text{pred}}\textbf{x}^{\text{new}}}}$ = \textproc{TrajOpt}($\textbf{x}^{\text{pred}}_L$, $\textbf{x}^{\text{new}}_L$) \label{lineinsat:trajopt}
\State $\>$ $\>$ $\>$ \textbf{if} isValid($\boldsymbol{\phi_{\textbf{x}^{\text{pred}}\textbf{x}^{\text{new}}}}$) \textbf{then}
\State $\>$ $\>$ $\>$ $\>$  $\boldsymbol{\phi_{\textbf{x}^{S}\textbf{x}^{\text{new}}}}$ = \textproc{WarmStartOpt} \label{lineinsat:warmopt}
($\boldsymbol{\phi_{\textbf{x}^{S}\textbf{x}^{\text{pred}}}}$, $\boldsymbol{\phi_{\textbf{x}^{\text{pred}}\textbf{x}^{\text{new}}}}$) 
\State $\>$ $\>$ $\>$ $\>$ \textbf{if} isValid($\boldsymbol{\phi_{\textbf{x}^{S}\textbf{x}^{\text{new}}}}$) $\cup$ c($\boldsymbol{\phi_{\textbf{x}^{S}\textbf{x}^{\text{new}}}}$) $<$ c($\textbf{x}^{\text{new}}_L$) \textbf{then}
\State $\>$ $\>$ $\>$ $\>$ $\>$ c($\textbf{x}^{\text{new}}_L$) = c($\boldsymbol{\phi_{\textbf{x}^{S}\textbf{x}^{\text{new}}}}$)
\State $\>$ $\>$ $\>$ $\>$ $\>$ Set $\textbf{x}^{\text{pred}}_L$ as the parent of $\textbf{x}_L^{\text{new}}$
\State $\>$ $\>$ $\>$ $\>$ $\>$ Store $\boldsymbol{\phi_{\textbf{x}^{S}\textbf{x}^{\text{new}}}}$
\State \textbf{return} $\boldsymbol{\phi}_{\textbf{x}^{\text{S}}\textbf{x}^{\text{G}}}$
\EndProcedure
\end{algorithmic}
\caption{INSAT}
\label{alginsat:insat}
\end{footnotesize}
\end{algorithm}

\vspace{-.3cm}

\subsection{Kinodynamic Planning using INSAT}
\label{sec:INSAT}
Kinodynamic planning is a class of problems for which velocity, acceleration, and inertial/force/torque bounds must be satisfied, together with kinematic constraints such as avoiding obstacles. However, controllers in fully actuated systems like the vast majority of commercial manipulators do not require a fully dynamically feasible trajectory to track them accurately. In such systems, even a velocity controller is able to track trajectories generated with smooth splines \cite{smoothmaniptraj} at high accuracy. As a result, planning in the space of the manipulator's joint space and its derivatives simplifies into finding the parameters of the choice of basis splines. This dramatic reduction in the planning complexity resulting from the spline representation of the manipulator trajectory enables us to use a recent global kinodynamic planning algorithm called INSAT \cite{insat1} as the preprocessing planner. In this section, we will first explain our choice of splines and provide a high-level overview of INSAT. We refer the reader to \cite{insat1, insat2, troptc} for the details of the algorithm and \cite{pinsat} for the parallelized version named PINSAT.

\subsubsection{B-Splines}
\label{sec:B-Splines}
B-splines are smooth and continuous piecewise polynomial functions made of finitely many basis polynomials called B-spline bases. A $k$-th degree B-spline basis with $m$ control points can be calculated using the Cox-de Boor recursion formula \cite{bsplines} as
\begin{equation}
N_{i, k}(t)=\frac{t-t_i}{t_{i+k-1}-t_i} N_{i, k-1}(t)+\frac{t_{i+k}-t}{t_{i+k}-t_{i+1}} N_{i+1, k-1}(t)
\end{equation}
where $i=0, \ldots, m$, $\frac{t-t_i}{t_{i+k-1}-t_i}$ and $\frac{t_{i+k}-t}{t_{i+k}-t_{i+1}}$ are the interpolating coefficients between $t_i$ and $t_{i+k}$. 
Let us define a non-decreasing knot vector $\textbf{T}$ 
and the set of control points $\textbf{P} = \textbf{P} = \{\textbf{p}_1, \textbf{p}_2, \ldots, \textbf{p}_m\}$ 
where $\textbf{p}_i \in \mathbb{R}^n, i=0,\ldots, m$. Then a B-spline trajectory can be uniquely determined by the degree of the polynomial $k$, the knot vector $\textbf{T}$, and the set of control points $\textbf{P}$ called de Boor points.
\begin{equation}
    \textbf{q}(t) = \sum_{i=0}^m \textbf{p}_i N_{i, k}(t) 
\end{equation}

The pseudocode of INSAT is given in Alg. \ref{alginsat:insat}. INSAT alternates between searching a low-D discrete graph and performing trajectory optimization in high-D to produce smooth full-D trajectories. The low-D variables in the algorithm are denoted with subscript $L$ and the full-D trajectories connecting two lifted spaces of low-D state are $\textbf{x}$ and $\textbf{x}^\prime$ is denoted as $\phi_{\textbf{x}\textbf{x}^\prime}$. For the \textproc{TrajOpt} step we solve the following optimization problem in Eq. \ref{eq:trajopt} using the aforementioned B-spline representation. The finite parameterization of this problem and how its optimization is interleaved with the low-D search is explained in high detail in \cite{pinsat}.

\begin{subequations}
\begin{align}
\begin{split}
\min \quad w_1 t_f + w_2 \int_{0}^{t_f} \mid \dot{\textbf{x}}(t) \mid_2 dt
\end{split}
\label{eq:trajoptobj} 
\\
\text{s.t.}\qquad & \textbf{x} \in \mathcal{C}^n \label{eq:constraint1} \\
& \textbf{x}(t) \in \mathcal{X}^{free} \label{eq:constraint2} \\
& \dot{\textbf{x}}(t) \in [\dot{\textbf{x}}_{min}, \dot{\textbf{x}}_{max}] \label{eq:constraint3} \\
& t_f \in [t_{min}, t_{max}] \label{eq:constraint4} \\
& \textbf{x}(t_0) = \textbf{x}_0, \textbf{x}(t_f) = \textbf{x}_f \label{eq:constraint5} \\
& \dot{\textbf{x}}(t_0) = \dot{\textbf{x}}_0, \dot{\textbf{x}}(t_f) = \dot{\textbf{x}}_f \label{eq:constraint5} 
\end{align}
\label{eq:trajopt}
\end{subequations}

The optimization is performed on the control points of the B-splines. The precise details of the optimization are not mentioned here due to the page limit. Once the preprocessing module's request is received, INSAT kicks in to find a smooth B-spline from start to goal. When the optimization output is valid in terms of dynamic limits but results in a collision, we recover the trajectory by caching the iterates and returning the best trajectory to the preprocessor \cite{pinsat}.

\subsection{Perception Module}
\label{sec:perception}
Predicting the trajectory of the incoming projectile is crucial for a successful intercept. This problem introduces two challenges:
\begin{enumerate}
    \item A single stereo camera is placed on the 8020 aluminum T-slotted profiles, mounted on the robot arm's pedestal. Consequently, precise camera calibration is important to establish the extrinsic rigid body transformation between the camera frame and the robot's base frame.

    \item The perception system should detect the rapidly moving projectile and provide an accurate projectile estimate of the projectile in real-time during its flight.
\end{enumerate}

We solve the first challenge with hand-eye camera calibration (\ref{sec:depth}) and solve the second challenge by performing a least-squares model fitting based on all observations (\ref{sec:lsf}).

\subsubsection{Object Detection}
To detect the incoming projectile, we employ RGB color filtering, as the projectile is characterized by a distinct colored ball. We establish specific lower and upper color thresholds within the HSV color space and then process each frame to generate a binary mask that isolates the ball.



\subsubsection{Depth Estimation}
\label{sec:depth}
To obtain the 3D global coordinates relative to the robot's base, we calibrate the camera to obtain the transformation between the left camera frame and the robot's base frame. Masked point cloud information and the confidence map are retrieved from the ZED SDK, with our Python API acting as a wrapper around the SDK. We filter outliers and eliminate pointclouds outside the predefined virtual bounding box. Then, we compute the mean values for $X$, $Y$, and $Z$ to derive the 3D centroid point. Only frames that exceed a minimum pixel threshold are considered to reduce noise. Since the ZED camera does not directly measure the depth but estimates it using stereo geometry, some depth points may not be entirely accurate. To address this, we utilize ZED SDK's built-in confident map function and drop depth points with low confidence. Consequently, we perform depth filtering, only retaining frames where the estimated depth falls within the specified distance threshold. The positions in these frames are recorded along with the timestamps, and this iterative process continues until the required number of frames is collected.\label{sec:depth} 
\subsubsection{Projectile Estimation}
\label{sec:lsf}

To reconstruct the 3D trajectory of the ball, we use the Ohno method \cite{Ohno2000TrackingPA}. First, we fit multiple position detections at various time intervals to a projectile equation of motion. It is assumed that the motion is on the $X-Z$ plane (with $Z$ pointing vertically upward). Movements perpendicular to this plane that can occur as a result of wind or other external forces are ignored. Once launched, the motion of the ball is governed purely by gravity, which is expressed as

\begin{gather}
    X(t)=X(0)+tV_{X}(0)\label{SUVAT1}\\
    Y(t)=Y(0)+tV_{Y}(0)\label{SUVAT2}\\
    Z(t)=Z(0)+tV_{Z}(0)-\frac{1}{2}gt^2\label{SUVAT3}
\end{gather}
where $(X(t), Y(t), Z(t))$ and $(V_{X}(t), V_{Y}(t), V_{Z}(t))$ denote the position and velocity of the ball at time $t$, $g$ denotes acceleration due to gravity ($g = 9.81 m/s^2$). Assuming parabolic motion for the projectile, the estimated position at time $t$ depends only on the initial position $(X(0), Y(0), Z(0))$ and the initial velocity $(V_{X}(0), V_{Y}(0), V_{Z}(0))$. We can then estimate the values of initial position and initial velocity $\theta = (X(0), Y(0), Z(0), V_{X}(0), V_{Y}(0), V_{Z}(0))$ by defining an error function that minimizes the sum of squared errors:
\begin{equation}
    \theta^*=\arg \min_{\theta} \sum_{t=1}^{M} \{(X_t-X(t))^2+(Y_t-Y(t))^2+(Z_t-Z(t))^2\}
\end{equation}
where $X(t),Y(t),Z(t)$ are the estimated positions from Eq. \ref{SUVAT1}-\ref{SUVAT3} and $X_t, Y_t, Z_t$ are observed positions from the detected frames in \ref{sec:depth}.



\section{Experiments and Results}
\label{sec:experiments}
\subsection{Evaluation in Simulation}
\begin{figure}[!htb]
    \centering
    \begin{tabular}{ccc}
        \includegraphics[width=0.14\textwidth, height=0.9in]{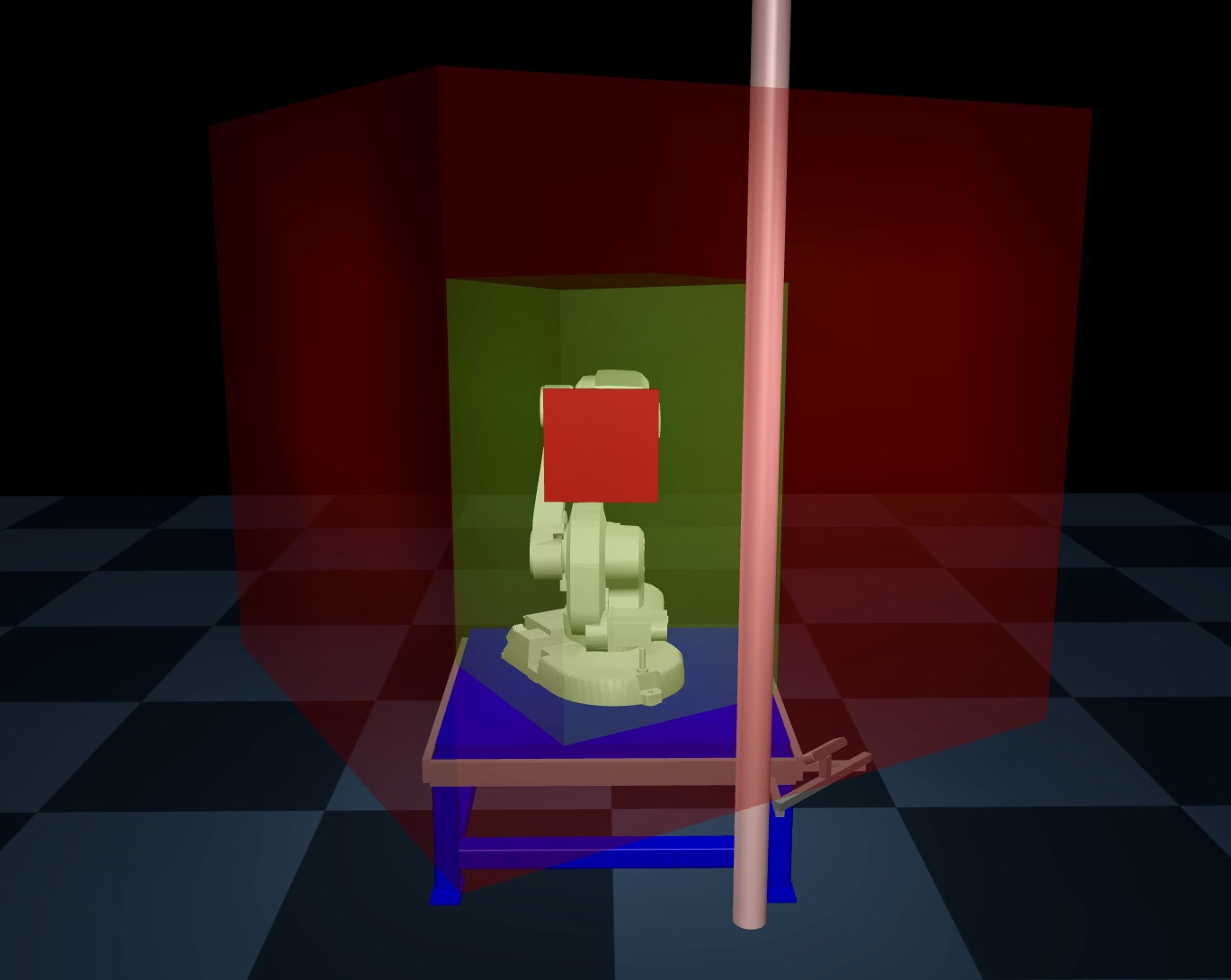} &
        \includegraphics[width=0.14\textwidth, height=0.9in]{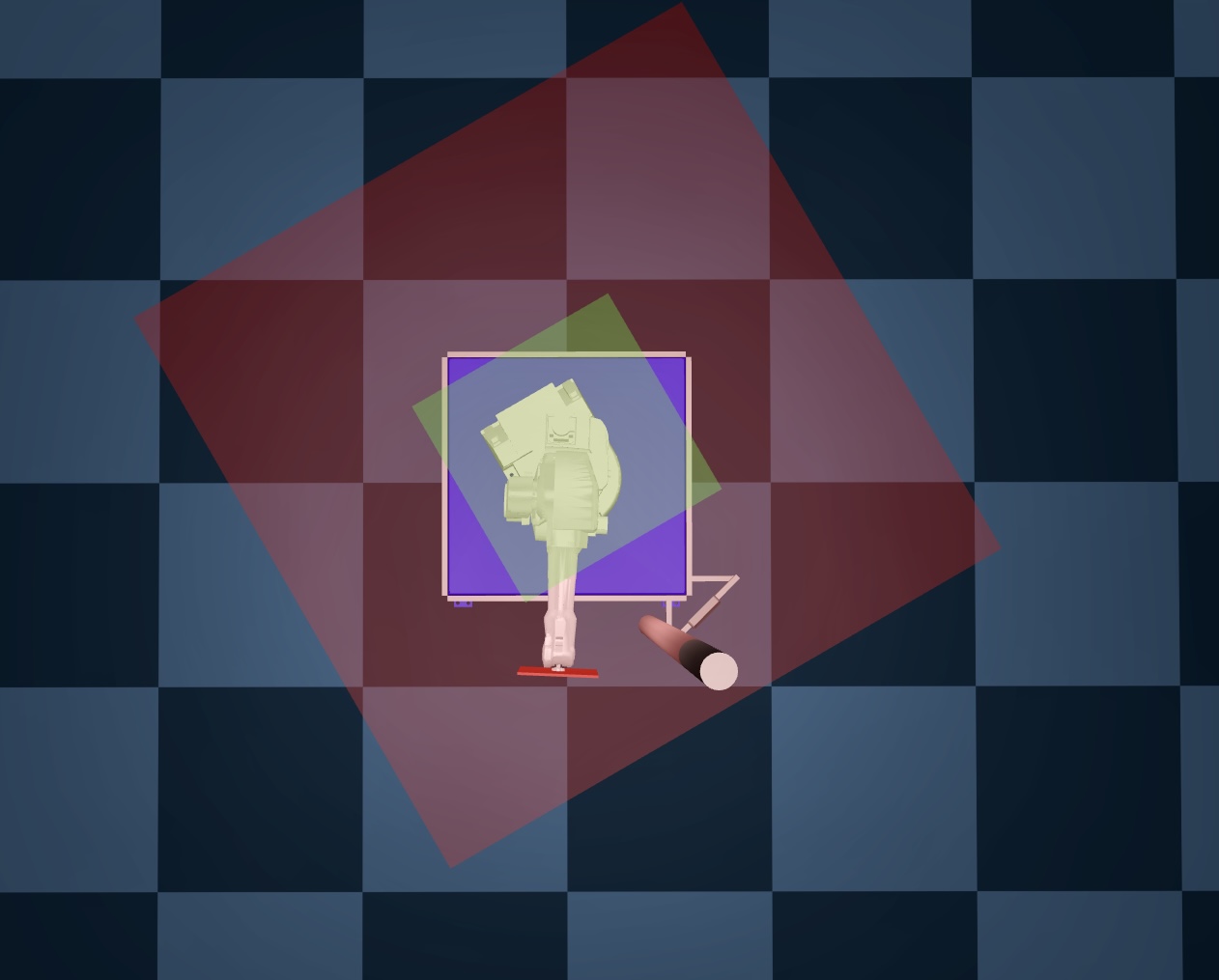} &
        \includegraphics[width=0.14\textwidth, height=0.9in]{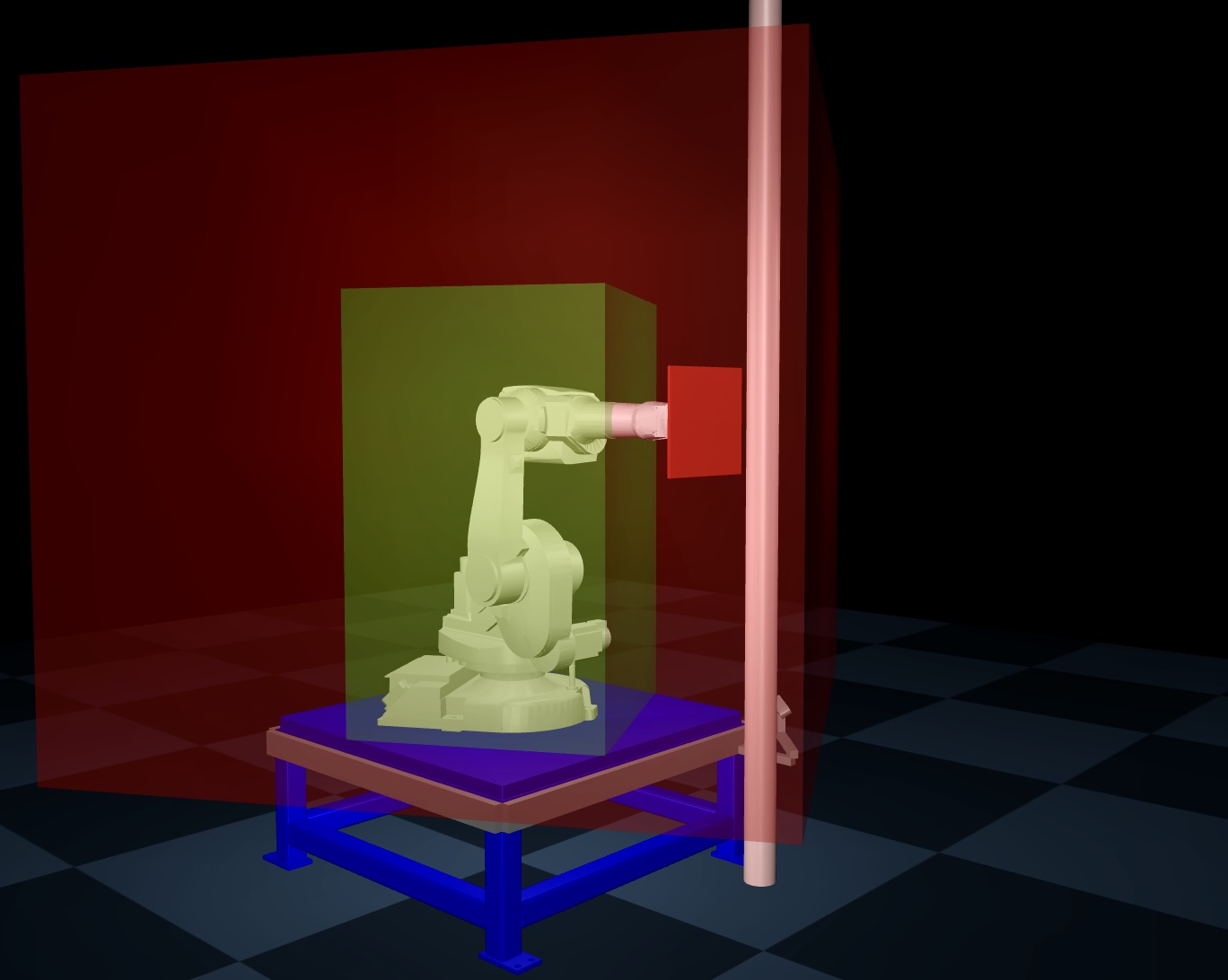} \\ 
\footnotesize{(a) Front view}
\label{fig:abb_front} &
\footnotesize{(b) Top view} 
\label{fig:abb_top}&
\footnotesize{(c) Side view} 
\label{fig:abb_top} \\
        \end{tabular}
        \caption{\footnotesize{(a), (b), and (c) are the front, top, and side views of the system in the Mujoco simulator.}}
        \label{abb_main}
        \vspace{-2mm}
\label{fig:experiment}
\end{figure}

We first evaluate the motion planning module in simulation. 
 The simulation environment configuration is shown in Fig. \ref{fig:experiment}, where the yellow cube denotes the inner dome and the red cube denotes the outer dome. A pink cylindrical pole is an obstacle in the environment. We randomly generated 200 projectiles for the experiment and verified that they intersect with the inner dome and outer dome. The launch distance for each projectile was uniformly sampled from a range of 6--12m. These projectiles have an average time of flight of 1.069s. Note that the time constraint is relaxed due to the addition of the obstacle, forcing the manipulator to take longer to execute a trajectory. All experiments were carried out on an AMD Threadripper Pro 5995WX workstation. We compared our planning framework with two efficient online planners: the Rapidly-exploring Random Tree-Connect (RRT-Connect) \cite{kuffner2000rrt} and the Edge-based Parallel A* (ePA*SE) \cite{mukherjee2022epa}. The RRT-Connect is known to be efficient in solving single-query path planning problems, while ePA*SE is a recent work that leverages the power of parallelization to speed up the search. All online planners were given a 2-second budget to plan. We used Toppra \cite{toppra} to generate a time-optimal trajectory from the geometric path from online planners. The preprocessing time of our planning framework is 5 hours.

\begin{table}[h]
\centering
\begin{tabular}{|l|l|l|l|}
\hline
\label{tab:plan_comp}
\textbf{Metrics}                                                                   & \textbf{\begin{tabular}[c]{@{}l@{}}Our approach \end{tabular}} & \textbf{\begin{tabular}[c]{@{}l@{}}RRT-Connect \\ \end{tabular}}   &\textbf{\begin{tabular}[c]{@{}l@{}}ePA*SE \\ \end{tabular}} \\ \hline
\begin{tabular}[c]{@{}l@{}} Finds a solution  \end{tabular}     & 70.33\%                                                                       & 59.81\%        & 50.71\%                                                                                   \\ \hline
\begin{tabular}[c]{@{}l@{}} Successfully Blocks \end{tabular}   & 68.9\%                                                                        & 27.27(27.27)\%           & 9.57(11)\%                                                                                \\ \hline
\begin{tabular}[c]{@{}l@{}} Query time (ms)\end{tabular}   & $0.109 \pm 0.033$                                                         & $17.5 \pm 31.7$     & $231 \pm 377$                                                                                         \\ \hline
\begin{tabular}[c]{@{}l@{}} Execution time (s) \end{tabular}   & $0.31 \pm 0.2$                                                 & $1.3 \pm 0.97$             & $1.8 \pm 0.87$                                                                              \\ \hline
\end{tabular}
\caption{Proposed method \textit{vs} the baselines in simulation.}
\label{tab:sim_result}
\vspace{-0.4cm}
\end{table}

The simulation results for 200 projectiles are shown in Table \ref{tab:sim_result}. The first row represents the ratio of successful interception solutions found by the planner. The second row indicates the ratio of successful blocking instances, defined as cases where the combined planning query time and execution time are shorter than the time of flight of the incoming projectile. The third and fourth rows present the mean and standard deviation of the planner query time and trajectory execution time. The simulation results demonstrate the superior performance of our method compared to the two baseline approaches. Specifically, our method achieves an exceptional success rate, nearly 2.5x of RRT-Connect and 7x that of ePA*SE. In terms of the query time, our method is faster than RRT-Connect by a factor of 160 and ePA*SE by 2119, affirming our planner's ability to swiftly compute manipulator trajectory paths within a short time window. Moreover, the execution time is substantially shorter than that of the two baselines, ensuring that the shield reaches the goal configuration in time for successful interception. Since the baselines are not constant time planners, we also computed the successful interception rate assuming zero query time for them (shown within parentheses in the second row). Table \ref{tab:torque} shows the joint-wise RMS torque in Nm across the simulation experiments. Notice a significantly lower net torque in our method for every joint. This is a direct benefit of interleaving search and optimization as opposed to post-processing geometric paths with path parameterizations subject to certain forms of kinematic and dynamic constraints.


\begin{table}[ht]
\renewcommand{\arraystretch}{1.5}
\centering
\resizebox{\columnwidth}{!}{%
\begin{tabular}{p{1.2cm}|cccccc}\toprule
\textbf{Joint ID} &  \textbf{1} & \textbf{2} & \textbf{3} & \textbf{4} & \textbf{5} & \textbf{6} \\ \hline\hline
\textbf{RRT-C} & 102 $\pm$ 40 & 130.2 $\pm$ 49 & 43.6 $\pm$ 24 & 0.35 $\pm$ 1.3 & 0.13 $\pm$ 0.35 & 0.015 $\pm$ 0.12 \\\hline
\textbf{ePA*SE} & 61.4 $\pm$ 38 & 131 $\pm$ 52 & 39.5 $\pm$ 14 & 0.34 $\pm$ 1.3 & 0.11 $\pm$ 0.25 & 0.012 $\pm$ 0.09 \\\hline
\textbf{Ours} & 11 $\pm$ 11 & 24.5 $\pm$ 21 & 16.2  $\pm$ 16 & 0.031 $\pm$ 0.02 & 0.036 $\pm$ 0.03 & 0.0003 $\pm$ 0.0003 \\
\bottomrule
\end{tabular}
}
\caption{Joint-wise RMS torque (Nm) for simulation experiments.}
\label{tab:torque}
\vspace{-.8cm}
\end{table}

\subsection{Experiments on Robot Hardware without Obstacles}

We tested the full system with integrated perception on the ABB's IRB-1600 robot arm, equipped with an onboard stereo camera ZED 2i, in an indoor environment. We configured the range of attacks to be 6--8m. In this setup, the time of flight of the ball from its first detection by the perception system to its interception with the robot is roughly 350ms. Of this time, on average, 82.9ms is allocated to the perception system, 212.2ms is dedicated to the execution of the robot's trajectory, and 50ms is attributed to other system overheads. This allocation leaves only 4.9ms on average for the planner. Nonetheless, our planner successfully generates a plan within this time budget.

We performed 50 throws and achieved a blocking success rate of 78\%. Among the 11 throws that were unsuccessful, the causes of these failures were traced back to issues within the perception and execution modules. Specifically, in 8 instances, inaccuracies in the perception module arose due to motion blur or noisy depth estimations. The remaining three failures were attributed to slow execution processes. This occurred when the execution time for specific trajectories either outpaced the perception module's ability to provide timely updates or when executing the entire trajectory consumed more time than the actual flight time of the ball.

\section{Conclusion and Future Work}
\label{sec:conclusion}
In this paper, we have presented and evaluated a preprocessing-based kinodynamic motion planning framework to intercept projectiles using a robot manipulator. We tested our overall pipeline, which consists of a perception module, a planner module, and an execution module on a physical system made of an ABB industrial arm and a ZED stereo camera. In the future, we would like to extend this work to a mobile base and intercept a sequence of projectiles separated by a short time interval.


\section{Acknowledgements}
\label{sec:acknowledgements}
This work was supported by grants W911NF-18-2-0218 and W911NF-21-1-0050 of the ARL-sponsored A2I2 program.

\bibliographystyle{ieeetr}
\bibliography{refs}

\end{document}